\documentclass{article}

\usepackage{todonotes}

\usepackage[final]{nips_2018}

\usepackage[utf8]{inputenc} 
\usepackage[T1]{fontenc}    
\usepackage{hyperref}       
\usepackage{url}            
\usepackage{booktabs}       
\usepackage{amsfonts}       
\usepackage{amssymb}
\usepackage{amsmath}
\usepackage{nicefrac}       
\usepackage{microtype}      
\usepackage{subfig}
\usepackage{graphicx}
\usepackage[font=footnotesize]{caption}

\title{Semantically-aware population health risk analyses}

\author{
  Alexander New, Sabbir M. Rashid, John S. Erickson, \\\textbf{Deborah L. McGuinness, and Kristin P. Bennett} \\
  Rensselaer Polytechnic Institute\\
  Troy, NY 12180 \\
  \texttt{\{newa, rashis2, erickj4, mcguid, bennek\}@rpi.edu} \\
}

\begin{document}

\maketitle

\begin{abstract}
One primary task of population health analysis is the identification of risk factors that, for some subpopulation, have a significant association with some 
health condition. Examples include finding lifestyle factors associated with chronic diseases and finding genetic mutations associated with diseases in precision health.  We develop a combined semantic and machine learning system that uses a health risk ontology and knowledge graph (KG) to dynamically discover risk factors and their associated subpopulations. Semantics and the novel supervised cadre model make our system explainable. Future population health studies are easily performed and documented with provenance by specifying additional input and output KG cartridges.  
\end{abstract}

\section[Introduction]{Introduction}

Population health studies discover risk factors associated with 
diseases within subpopulations.
We found thousands of published studies that used multivariate logistic regression to identify risk factors using the National Health and Nutrition Examination Survey (NHANES) dataset\footnote{https://www.cdc.gov/nchs/nhanes/index.htm},
yielding valuable population health insights at a great duplication of effort.\footnote{On 10/23/18, scholar.google.com returned 5240 results for the search ``NHANES `multivariate logistic regression' `risk factors'''.}
For each paper, the author typically specifies some combination of a single disease, a small number of risk factors, and  a specific patient cohort (e.g., a subpopulation defined by age, gender, or ethnicity).  Started in 1960, NHANES examines about 5000 subjects a year and thus serves as one of the primary data resources for epidemiology and population health studies.

In these studies, the goal is not to develop a supervised model with a high predictive accuracy, but rather to find significant associations between risk factors (e.g. exercise, nutrition, exposure to potential toxins) and diseases (e.g. cancer, cardiovascular disease, diabetes, vision loss) in a defined subpopulation.  Once validated in further studies, these risk factors can serve as the basis for interventions at the clinical and/or public policy levels leading to health promotion and disease prevention. 

By leveraging a combination of semantics and analytics, we can formalize and generalize these workflows to enable future studies to be performed dynamically with minimal effort. We utilize the environment-wide association study (EWAS, \cite{Patel2010}) and make it applicable to a wide array of 
diseases and risk factors by simply specifying high-level ``cartridges" that encode the study definition of interest.
The approach is coupled with interpretable precision health models that perform automatic identification of at-risk subpopulations with distinct risk profiles.  We emphasize that these subpopulations and risk profiles are discovered, which contrasts with the more typical study in which the researcher designs a cohort and then runs a targeted analysis of a few risk factors.

Our risk analysis tool performs dynamic, 
semantically-aware precision risk analyses by testing the association between risk factors and health conditions or diseases for statistical significance. Any risk factor, health condition or disease definable in NHANES may be analyzed. As NHANES is updated, our tool can empower and accelerate novel population health studies and findings. In NHANES, data collection is biased to better capture minority populations; thus, we employ survey-weighted approaches (\cite{SurveyBook}) to correct for this bias.  Currently, we support survey-weighted generalized linear models with and without subpopulation discovery. The proposed approach can be rapidly adapted to alternative datasets and different analysis pipelines.    

The Risk Analyzer web application leverages our ontology and a knowledge graph (KG) to drive analytics performed in R. 
As described in (\cite{McCusker2018}), a KG is a graph in which meaning is expressed as structure, statements are unambiguous, a limited set of relationships are used, and explicit provenance is included.
We build on prior semantic representations of analytics, including the foundational semantic workflow model (\cite{McGuinness2015,Bennett2016}) and the ScalaTion (\cite{Nural2015}) framework. Underlying everything is a representation of data and metadata in our KG. The KG uses best practices ontologies and the Cohort Ontology (\cite{CohortOntology}) extended to capture risk factor analyses using ``cartridges". The derived subpopulations and risk profiles are also declaratively encoded in the KG.

\section{Methods}

\subsection{Preparation}\label{sec:preparation}

We first ingest NHANES records into a linked data representation with the Semantic Data Dictionary (SDD, \cite{rashidsemantic2017}) approach. 
This creates 
a dictionary mapping and a codebook for NHANES variable values, cohorts, and survey weights.  Both of these link NHANES
terminology to terms in open source ontologies that support reuse and interoperability.
Ontologies that we reuse include the Semanticscience Integrated Ontology (SIO), Children's Health Exposure Analysis Resource (CHEAR), Statistics Ontology (STATO), and Chemical Entities of Biological Interest (ChEBI).  This allows for the creation of a KG of data and metadata, which can be 
interlinked 
with best practice terminology. Further, the content is published using nanopublications (\cite{Groth2010}) to maintain provenance.
We develop a Health Risk Factor Ontology
that contains three 
classes of entities: analysis concepts, workflow axioms, and background domain axioms. 
Analysis concepts include 
definitions of responses and associated confounding variables.
Workflow axioms describe data preprocessing steps, such as standardization of continuous risk factors. 
Background domain axioms capture knowledge necessary for specifying the analytics model. 
For example, to make urine measurements of risk factors comparable, the analysis can control for urinary creatinine. 

Minimal necessary analysis components from the Health Risk Factor Ontology are stored in modular serializations called cartridges, and these are loaded as the user chooses how to construct their study. Response variable cartridges provide the analysis concepts and background domain axioms necessary to model a given disease. Study cohort cartridges list inclusion criteria used to determine if a given subject 
may be included in the study. Risk factor cartridges describe categories of semantically similar risk factors and give their necessary background domain axioms. Workflow cartridges contain a set of axioms that completely detail an analysis workflow. Once an analysis has been performed, the findings are stored in a results cartridge, which also contains links to the other cartridges used to perform that analysis. Results are written back to the KG so that they can be accessed by query.

\subsection{Analysis}

The user designs their study using the Risk Analyzer's graphical interface with defaults informed by the KG. This requires choosing the target chronic health condition, control factors, risk factors, and patient cohort. One study might examine 
associations in adult patients between 
type 2 diabetes (T2D) and a hundred different pesticides while controlling for age, gender, ethnicity, and BMI. Based on these settings, the Risk Analyzer queries the KG using SPARQL (\cite{harris2013sparql})
to extract the relevant records. Then it applies the appropriate workflow axioms and visually presents the results.

The user may choose different survey-weighted analysis methods, including generalized linear models for population analysis and the supervised cadre model (SCM, \cite{New2018}) for interpretable precision health. Cadre models simultaneously discover subpopulations and learn their risk models. Subpopulations, which we call cadres, are subsets of the population defined with respect to a rule learned by the SCM. Subjects in the same cadre have the same association with a given risk factor. 

SCMs may be applied to a variety of analytics tasks, including multiple and multivariate regression and classification. We sketch out binary classification. When trained on a set of observations $\{x^n\}\subseteq \mathbb{R}^P$, the supervised cadre models divides the observations into a set of $M$ cadres. Each cadre $m$ is characterized by a center $c^m \in \mathbb{R}^P$ and a linear risk function $e_m$ parameterized by $w^m \in \mathbb{R}^P$. New observations $x$ have a aggregate risk score given by $f(x) = \sum_{m=1}^M g_m(x) e_m(x)$, where
\[g_m(x) = \frac{e^{-\gamma ||x - c^m||_d^2}}{\sum_{m'}e^{-\gamma ||x - c^{m'}||^2_d}}\,\,\,\,\,\,\,\,\mathrm{and}\,\,\,\,\,\,\,\,e_m(x) = \left(w^m\right)^T x.\]
Here $g_m(x)$ is the probability $x$ belongs to cadre $m$, $||z||_d = \left(\sum_p |d_p| (z_p)^2\right)^{1/2}$ is a seminorm parameterized by $d\in \mathbb{R}^P$, and $\gamma > 0$ is a hyperparameter that controls the sharpness of the cadre-assignment process. The cadre membership of $x$ is a multinoulli random variable with probabilities $\{g_1(x),\hdots,g_M(x)\}$; this set is the softmax of the set of weighted inverse-distances $\{\gamma||x-c^1||_d^{-2},\hdots,\gamma||x-c^M||_d^{-2}\}$. The optimal parameters for an SCM are obtained by applying stochastic gradient descent to a loss function derived from a posterior maximization problem.

The supervised cadre model is ante-hoc explainable (\cite{Holzinger2018}): it incorporates explainability directly into its structure. 
Its parameters have intuitive purposes. The feature distributions of observations can be grouped by cadre to reveal underlying cadre structure. For example, an SCM might discover that, for the subpopulation of middle-aged and older white people, there is a significant association between the pesticide nicosulfuron and high systolic blood pressure (SBP).

After the analysis, new metadata and data are written back into the KG, 
annotated with terms defined in the ontology.  
The metadata include subpopulation summary statistics, such as variable means and standard deviations, and SCM hyperparameters. 
The data include each observation's cadre membership.
The ontology was designed to include attributes useful for informing analysis design, summarizing analysis results for later comparison, and for generating reports.
By including annotations that use well defined terms, we claim that the resulting knowledge graph can be considered a computational knowledge graph, as applications can compute with statements embedded in it since they have definitions of terms used and also have the provenance of how the statements were derived.


\section{Results}

\begin{figure}
    \centering
    \includegraphics[width=0.9\linewidth]{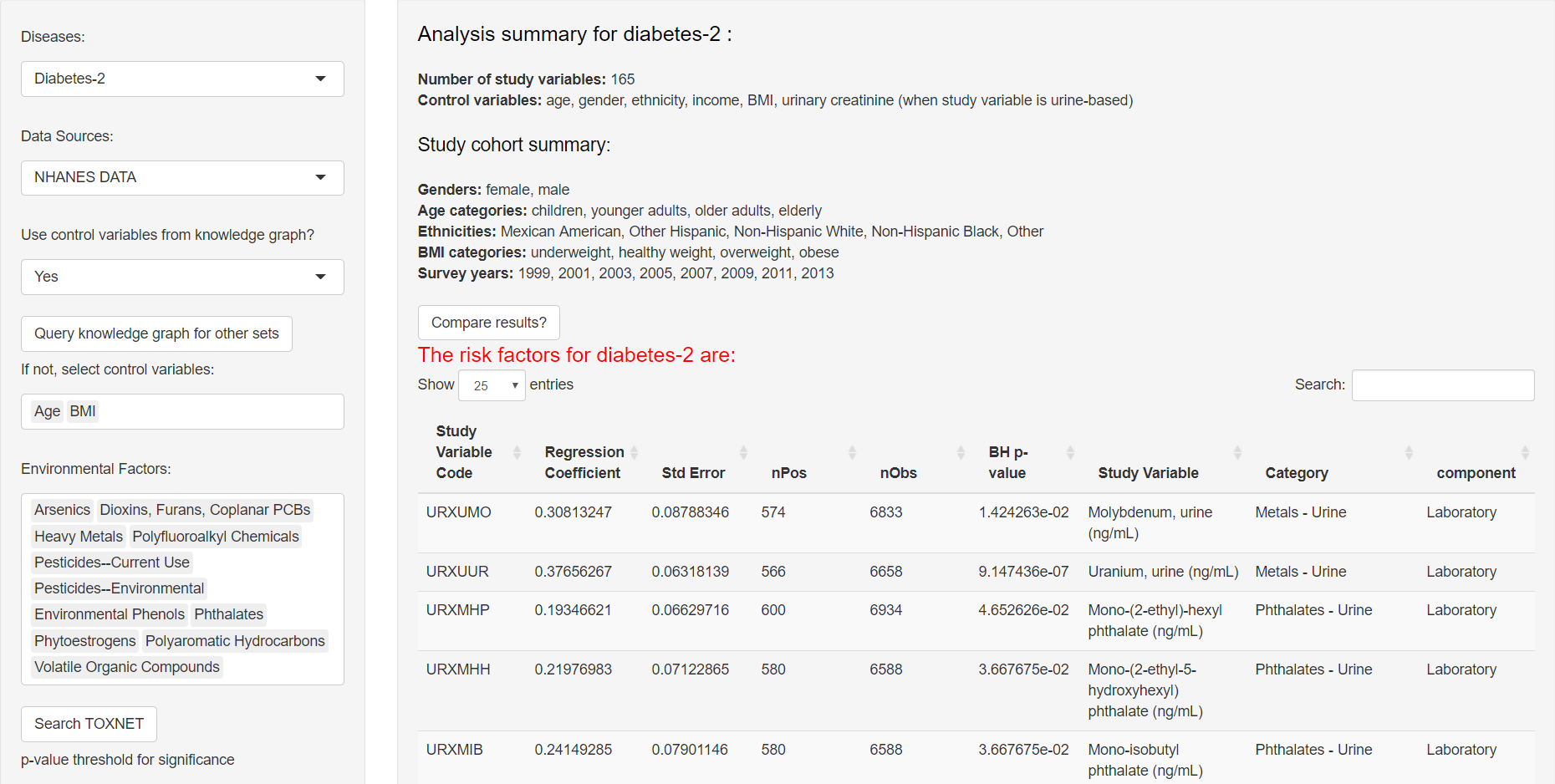}
    \caption{\footnotesize Dynamic study by Risk Analyzer of T2D risk factors, finding evidence urinary molybdenum and urinary uranium are associated with T2D. The KG drives an R-Shiny web application enabling the user to conduct customized risk analyses for T2D, heart disease, breast cancer, hypertension, and thyroid disease.}
    \label{fig:risk-analyzer}
\end{figure}

\begin{figure}
    \centering
    \includegraphics[width=\linewidth]{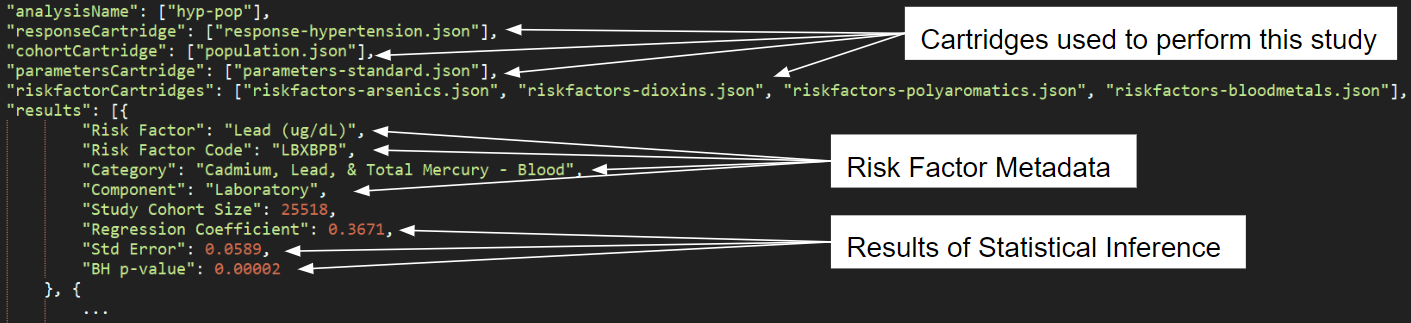}
    \caption{\footnotesize Results cartridge for study finding environmental factors associated with hypertension. Blood lead is significant, matching prior work by \cite{bloodLead1} and \cite{BloodLead2}.}
    \label{fig:cartridge}
\end{figure}

The risk analyzer analyzes a user-specified subpopulation with the Health Risk Factor Ontology.
Possible risk factors include 218 environmental exposure and 36 lifestyle variables. It targets 
diseases: T2D, breast cancer, heart disease, thyroid disease, and hypertension. Figure \ref{fig:risk-analyzer} shows the results of this implementation being used to identify environmental factors associated with type 2 diabetes. The user dynamically specifies and executes the study through the interface. Figure \ref{fig:cartridge} shows a portion of a results cartridge. This cartridge stores the findings that identified environmental factors associated with high SBP.

\begin{figure}
    \centering
    \subfloat[][Gender and ethnicity counts by cadre ]{\includegraphics[width=0.45\linewidth]{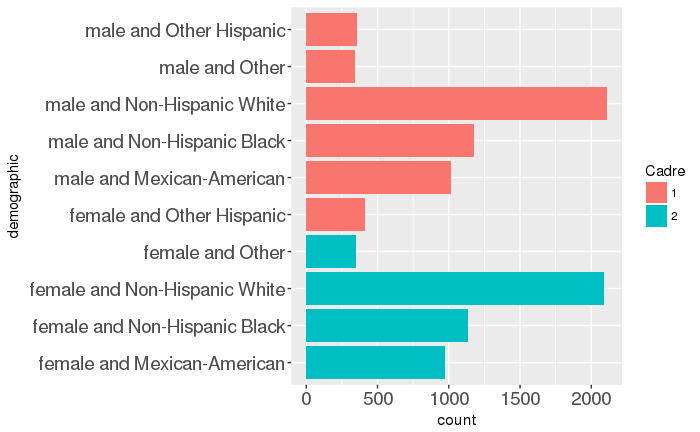}}
    \hfill
    \subfloat[][Variable distributions by cadre ]{\includegraphics[width=0.45\linewidth]{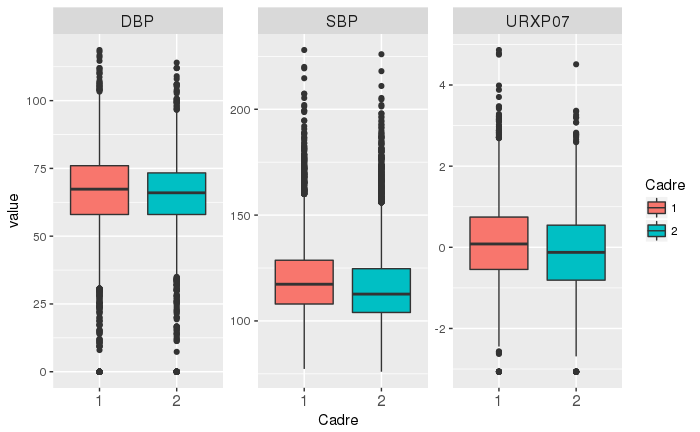}}
    \caption{\footnotesize Two subpopulations discovered by the SCM. In the first (Cadre 1), there is a significant association between 2-hydroxyphenanthrene (URXP07) and hypertension. In the second, the evidence was not sufficient to find a significant association. Cadre 1 consists of men and Other Hispanic women, and its members typically have higher diastolic and systolic blood pressures, as well as higher concentrations of URXP07 in their urine.}
    \label{fig:subpopulations}
\end{figure}

We can examine the subpopulations discovered by the supervised cadre model. Fig. \ref{fig:subpopulations} shows two subpopulations associated with the polyaromatic hydrocarbon 2-hydroxyphenanthrene. These subpopulations and their discovered associations can also be written back to the knowledge graph.

\section{Discussion}

We have presented a unique synthesis of semantics and analytics and used it to execute population health risk factor studies. 
We leverage inference capabilities and ontology term relationships to dynamically construct and execute the risk factor model and interpret the results. Using cadre models linked with semantics, the system can provide discovery and explainable insights in future population health studies. Our system writes its statistical findings and parameters back to the KG, enabling them to be retrieved via query while enabling comparisons with previous results. By linking a result to the cartridge used to produce it, we completely document how that result was produced.  

Future work includes a full implementation of the SCM framework and expansion of the Health Risk Factor Ontology, enabling users to dynamically introduce and apply cartridges modeling new diseases and risk factors of interest to different datasets.  Predicate relationships will facilitate the analytics.   We report here on population health analyses of survey data using cadre models. This same framework could readily be adapted to the analysis of electronic health care records to support evidence-based analyses targeted towards individual patients. Multimodal narrative generation and presentation technologies can use the KG to dynamically create interactive interfaces and reports that communicate and explain analytic insights targeted towards researchers, clinicians or patients.  



\subsubsection*{Acknowledgments}
This work is partially supported by IBM Research AI through the AI Horizons Network.


\bibliography{biblio}

\end{document}